\documentclass[letterpaper, 10 pt, conference]{ieeeconf}  

\usepackage{graphicx}
\usepackage{amsmath}
\usepackage{subfigure}
\usepackage{xcolor}
\usepackage{subcaption}
\IEEEoverridecommandlockouts                              

\title{\LARGE \bf
Haptic feedback of front car motion can improve driving control}

\author{Xiaoxiao Cheng$^{1,2*}$, Xianzhe Geng$^{1*}$,  Yanpei Huang$^{1,3}$ and Etienne Burdet$^1$ 
\thanks{* Equal contribution. This work was supported by the EC H2020 grants CONBOTS (ICT 871803) and NIMA (FETOPEN 899626).\newline
$^1$Department of Bioengineering, Imperial College of Science, Technology and Medicine, London, UK. 
$^2$Department of Electrical and Electronic Engineering, University of Manchester, Manchester, UK
$^3$School of Engineering and Informatics, University of Sussex, Brighton, UK.
{\{xiaoxiao.cheng@manchester.ac.uk, yanpei.huang@sussex.ac.uk, eburdet@ic.ac.uk\}}}
}

\begin{document}

\maketitle
\thispagestyle{empty}
\pagestyle{empty}

\begin{abstract}
This study investigates the role of haptic feedback in a car-following scenario, where information about the motion of the front vehicle is provided through a virtual elastic connection with it. Using a robotic interface in a simulated driving environment, we examined the impact of varying levels of such haptic feedback on the driver's ability to follow the road while avoiding obstacles. The results of an experiment with 15 subjects indicate that haptic feedback from the front car's motion can significantly improve driving control (i.e., reduce motion jerk and deviation from the road) and reduce mental load (evaluated via questionnaire). This suggests that haptic communication, as observed between physically interacting humans, can be leveraged to improve safety and efficiency in automated driving systems, warranting further testing in real driving scenarios.
\end{abstract}

\section{Introduction}

Driving on today's roadways is increasingly challenging due to persistent traffic congestion, complex road conditions, and the demanding nature of long journeys. Whether traveling on highways or navigating urban streets, car-following is a common driving task where drivers' actions are influenced by the vehicle in front of them. This task becomes particularly difficult in fluctuating traffic and complex road conditions. Recent statistics show that there are approximately two million rear-end collisions each year in the US, accounting for nearly 20 percent of fatalities in two-vehicle collisions \cite{nhtsa2023}.

Traditionally, drivers rely on their vision to gauge road conditions, primarily assessing factors such as the relative speed of the leading vehicle. However, visual perception has inherent delays \cite{vanrullen2001time}, which can be exacerbated by driver fatigue or distraction. Additionally, in car-following scenarios, a driver's view is often restricted by the vehicle in front, limiting their perception of other vehicles and the road ahead. In certain exceptional circumstances, such as abrupt speed changes of the vehicle ahead or the sudden appearance of obstacles, trailing drivers have minimal time to react. Complementing vision, haptic feedback has been shown to improve performance in physical human-human and human-robot collaboration by the spontaneous exchange of motion planning information between interacting partners \cite{hendrik2022physically}. This haptic communication could be useful for effectively transferring information in car-following situations, where interacting drivers have different visual information.

Haptic shared driving has been identified as a promising approach to facilitate driving and enhance safety \cite{abbink2012haptic}, e.g. by helping drivers to maintain lane centering \cite{wang2017effect, mars2014analysis, benloucif2019cooperative, wang2022modeling}. Shared-control methods have also been used to combine inputs from the human driver and the car's controller to optimize path-following and reduce workload \cite{wang2016human, li2020indirect}. Various authority transition schemes have been explored to improve smoothness during transitions between human drivers and automatic controllers \cite{cutlip2021effects, okada2020transferring, saito2018control}. Additionally, warnings and interventions have been implemented to prevent unintentional lane departures \cite{merah2016new, li2019shared}. These studies indicate that haptic feedback can provide beneficial driving guidance to human drivers. However, they have primarily focused on haptic shared driving within a single car, without considering surrounding vehicles and the potential for sharing their unique sensory information through haptic feedback.

To systematically investigate the influence of haptic feedback on driving performance and its potential to transmit sensory information between cars, we developed a haptic-shared driving simulator and conducted a user study in a virtual car-following scenario with random obstacles. An autonomous frontal car with consistent driving behavior was designed using a model predictive control (MPC) algorithm. The rear car was controlled by a human driver who received haptic feedback reflecting the motion of the frontal car. To evaluate the effects of haptic feedback under different driving conditions, we observed 15 subjects driving a virtual car at various velocities (10\,m/s, 12.5\,m/s, 15\,m/s) paired with different levels of haptic feedback (using stiffness values K\,=\,0, 200\,N/m, 500\,N/m). We analyzed the driving performance and workload of each subject based on evaluation metrics and questionnaires.

\section{Haptic-shared Driving System}
\subsection{Overview of the System}
\begin{figure}[t]
      \centering      
      \includegraphics[width=0.5\textwidth]{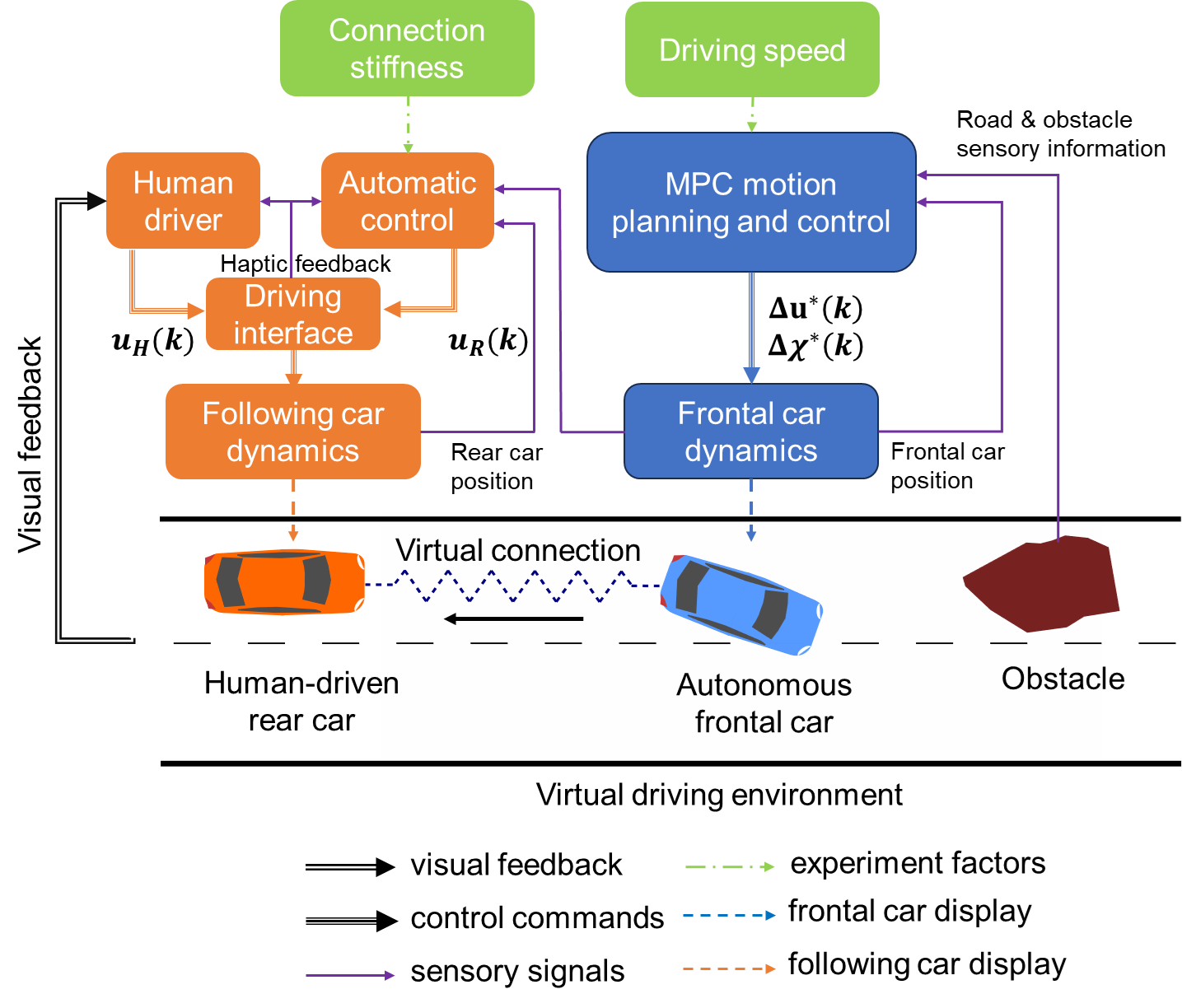}
      \caption{Architecture of the developed virtual driving system.}
      \label{setup}
\end{figure}

Our driving simulator features a haptic hand controller and a virtual car-following scenario with two vehicles (Fig.\,\ref{setup}). The lead car (blue) travels at a constant speed (selectable from 10\,m/s, 12.5\,m/s, or 15\,m/s) controlled by an MPC algorithm (detailed in Section \ref{s:frontcar}). This algorithm automates the car's movement, including obstacle avoidance, mimicking expert driving behavior. The following car (orange) is in cruise control mode, maintaining the same speed and a safe distance of 3.3 meters from the lead car.
 
Both cars drive on a simulated one-way, circular road with two lanes, each 3 meters wide, mirroring real road configurations. The total length of the loop is 408 meters and includes four turns, each greater than 90$^\circ$. The cars are modeled as 4.7x1.8x1.4\,m$^3$ cuboids. Five 2x2x0.9\,m$^3$ cuboid obstacles are randomly placed on either side of the road at intervals ranging from 20 to 40 meters. The cars can freely move across lanes to avoid these obstacles.
 
Haptic feedback to the driver of the rear car is provided by a H-Man robotic interface with a force range of [0, 7]\,N on a 340x340\,cm$^2$ horizontal workspace . The operator holds the handle and controls the rear virtual car's steering at 200\,Hz. The cars' movements are displayed on a monitor, providing visual feedback to the operator. Haptic feedback to the rear car's operator is based on the driving motion of the front car, as detailed in Section \ref{s:followcar}.

\subsection{Model Predictive Front Car Control} \label{s:frontcar}
To guarantee consistent haptic feedback, the lead car acts autonomously, controlled by an MPC algorithm. This algorithm plans the car's movements by anticipating future events and calculates optimal control actions for a set timeframe. It achieves this by optimizing a cost function that prioritizes goals like following a desired path and avoiding obstacles \cite{lazcano2021mpc}.

We firstly consider a 3 degree-of-freedom (DoF) kinematic car model \cite{matute2019experimental} to compute the car's Cartesian position $(x,y)$ and heading angle $\varphi$, where the control inputs are the current velocity $v$ and steering angle $\delta$:
\begin{align}
\begin{pmatrix} 
\dot{x}\\
\dot{y}\\
\dot{\varphi}
\end{pmatrix}
&=
\begin{pmatrix}
 v\cos \varphi \\
 v\sin \varphi \\
 \dfrac{v}{l} \tan \delta 
\end{pmatrix}
\label{1}
\end{align}
The control system can be represented as a state-space model $\Dot{\chi}=f(\chi,u)$ with control inputs $u=(\delta,a)^T$ and car state $\chi=(x,y,\varphi)^T$. 
Next, linearisation is carried out along the centre line of the road $(\chi_r,u_r)$:
\begin{align}
\begin{split}
\Dot{\chi}=f(\chi_{\text{r}},u_{\text{r}})&+\left.\frac{\partial{f(\chi,u)}}{\partial\chi}\right|_{\chi=\chi_{\text{r}},u=u_{\text{r}}}\!\!\!\!\!\!\!\!\!\!\!\!\!\!\!\!\!\!\!\!(\chi-\chi_{\text{r}})
+\left.\frac{\partial{f(\chi,u)}}{\partial{u}}\right|_{\chi=\chi_{\text{r}},u=u_{\text{r}}}\!\!\!\!\!\!\!\!\!\!\!\!\!\!\!\!\!\!\!\!(u-u_{\text{r}}) 
\end{split}
\label{eq_linerisation}
\end{align}
Let us denote the new state $\eta = \chi - \chi_r$ and control inputs $w = u-u_r$, representing the deviation of current car position from the reference trajectory. 
By descretising the linearised system (\ref{eq_linerisation}) using Euler method \cite{butcher2016numerical} with a time step $h$, the discrete state-space vehicle model yields:
\begin{align}\label{eq_discrete}
&\eta(k+1)= A_\eta(k)\,\eta(k) + B_\eta(k)\,w(k) \\
&A_\eta(k)  = M(k)\,h + I, \  B_\eta(k) = N(k)\,h \nonumber \\
&M(k)  = \frac{\partial{f(\chi,u)}}{\partial\chi}\bigg\rvert_{\chi=\chi_{\text{r}}(k),u=u_{\text{r}}(k)} \nonumber\\
&N(k) = \frac{\partial{f(\chi,u)}}{\partial{u}}\bigg\rvert_{\chi=\chi_{\text{r}}(k),u=u_{\text{r}}(k)} \nonumber
\end{align}
 
The car's next pose deviation is determined by the current pose deviation and the last control variables. Next, an extended system state by in cooperating the control input at the last time step is introduced:
\begin{align}
\xi(k+1)=\begin{pmatrix}\eta(k+1)\\w(k)
\end{pmatrix}
\label{5}
\end{align}
The extended system can be represented with the increment of the control input:
\begin{align} \label{eq_extendedSystem}
& \xi(k+1)= A_\xi(k)\,\xi(k) + B_\xi(k)\,\Delta w(k),\\
& A_\xi = \begin{pmatrix}
A_\eta & B_\eta\\0 & I    
\end{pmatrix}, \, B_\xi = \begin{pmatrix}
    B_\eta\\I
\end{pmatrix} \nonumber.
\end{align}
The MPC plans future states over the predictive horizon $N$. The future states at the future time step $i\,(1\le i\le N)$ can be calculated as
\begin{align}
\begin{split}
\xi(&k+i)=A_\xi^{i}\xi(k)+A_\xi^{i-1}B_\xi\Delta w(k)\\&+A_\xi^{i-2}B_\xi\Delta w(k+1)+\cdots+B_\xi\Delta w(k+i-1)
\end{split}
\label{7}
\end{align}

The following cost function is designed to ensure that the car closely follows the desired trajectory while avoiding obstacles:
\begin{align} 
\begin{split}
J&=\sum_{i=1}^{N}||\xi(k+i)||_{\text{Q}}^2+||\Delta w(k+i)||_{\text{R}}^2\\&+\sum_{i=1}^{N}\frac{Sv}{(x_i-x_{\text{c}})^2+(y_i-y_{\text{c}})^2+\zeta}
\end{split}
\label{eq_costfunc}
\end{align}
where $S$ is weighting coefficient, $(x_i,y_i)$ and $(x_{\text{c}},y_{\text{c}})$ are the positions of the car and the obstacle correspondingly. 

Given that the control objective is to minimize the deviation from the reference path, achieve obstacle avoidance and keep the car on the road, the specific form of the controller is given by
\begin{align}
\text{min}\ J, \text{s.t.}\begin{cases}
A\Delta w\le w_{\max}-w\\
A\Delta w\le -w_{\min}+w\\ 
\Delta w_{\min}\le \Delta w \le \Delta w_{\max}
\end{cases}
\label{9}
\end{align}

The MPC control inputs are obtained using the ACADO Toolkit \cite{lazcano2021mpc}, \cite{houska2011acado} by solving the defined optimisation problem. Obstacles and road boundaries detected within predictive horizons are considered as costs and constraints respectively, generating a collision-free trajectory at each time step for the predictive horizon.

\subsection{Haptic-shared Control of Rear Car} \label{s:followcar}
The control command for the front car is derived from the addition of the operator’s input and the force guidance from the automatic controller:
\begin{align}
u = u_{\text{human}} + u_{\text{feedback}}
\label{11}
\end{align}
The automatic controller calculates the steering command based on the car's current position as well as the force fed back to the H-Man handle, computed as:
\begin{align}
u_{\text{feedback}}=K_{\text{p}}(p-p^*)+K_{\text{d}}(\dot{p}-\dot{p}^*)
\label{10}
\end{align}
where $K_p$ and $K_d$ are the controller gains, $p, \dot{p}$ the current position and velocity of the rear car, $p^*,\dot{p}^*$ the position and velocity of the front car.
The human operator then applies a steering force on the handle, interacting with the automatic controller and adjusting the car's motion according to their assessment of the driving situation.

The virtual driving system was validated in simulations first through a driver model mimicking diverse driving abilities as can be seen in https://youtu.be/BVJoX7tLQEw.
\begin{figure*}[thpb]
\centering      
\includegraphics[width=1\textwidth]{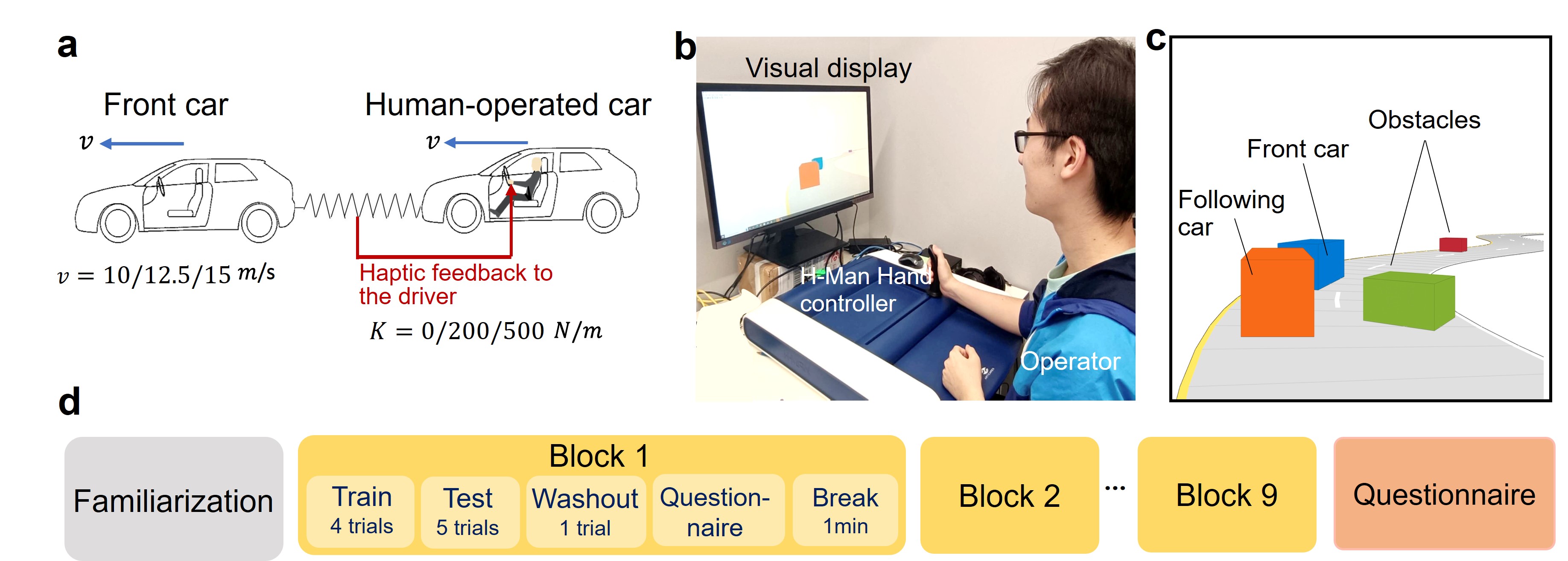}
\caption{Illustration of experiment protocol. a) Experiment conditions. b) Setup with H-Man haptic interface. c) Visual feedback from the driving simulator. d) Experiment sequence.}
\label{pro}
\end{figure*}

\section{Experiment}
An experiment was conducted to study the effect of haptic feedback passing information of the front car to the human drivers in the rear car. Human drivers in the rear car received varying levels of haptic feedback at different driving speeds. The experiment was approved by the Imperial College London ethics committee (No. 15IC2470). 15 subjects without motor impairment (12 male, 3 female; age = 22 $\pm$ 2 years) participated in the study. Nine out of the 15 participants have driving experience. 13 participants are right-handed (Edinburgh Handedness Inventory score $>$\,40 \cite{oldfield1971assessment}) and 2 ambidextrous participants. 


\subsection{Experimental Setup and Procedure}
The experiment setup is illustrated in Fig.\,\ref{pro}b-c. Each subject sits comfortably in front of the monitor on a height-adjustable chair and grasps the handle of the H-Man with their dominant hand. They received visual and haptic information on the virtual driving system in view from the far rear bumper camera \cite{salen2003rules}, which was designed to facilitate their perception of car steering angle (Fig.\,\ref{pro}c).

The participants were tasked with controlling the rear car to follow the road, avoid obstacles, and complete laps on a designed circular track during each trial. The experiment comprised nine conditions aimed at systematically studying the effects of shared haptic feedback. Haptic feedback was varied across three levels: no feedback, small feedback gain (controller stiffness of 200\,N/m), and large feedback gain (controller stiffness of 500\,N/m). In trials without feedback, the front and rear cars operated independently at a constant speed. In trials with small or large feedback gain, the motion feedback from the front car was transmitted to the rear car through a virtual spring mechanism. The feedback levels were chosen based on previous research \cite{ivanova2020motion} and practical considerations. Driving speed was set at three levels: \{10,12.5,15\}\,m/s aimed at simulating different driving conditions and varying operational challenges.

The experimental protocol is shown in Fig.\,\ref{pro}d. Participants first go through a \textit{familiarization phase}, during which the experimenter instructs them on how to use the system and perform driving operations. In the \textit{test phase}, participants navigate the rear car using the hand controller to follow the defined path and avoid obstacles. This phase includes nine blocks, each corresponding to a different combination of speed and haptic feedback conditions (a 3×3 two-factor factorial design). The sequence of conditions was randomized for each participant. Each block consists of ten trials: the first four are \textit{practice and training trials}, followed by five recorded \textit{task trials}, and a final washout trial before moving to the next block. At the end of each block, participants completed a NASA questionnaire to assess the workload associated with that specific block. To prevent fatigue, there was a one-minute break between consecutive blocks. After completing the test phase, participants filled out a final questionnaire on their preference for large versus small feedback gains and ranked the most significant factors influencing their driving performance, such as driving speed, haptic feedback, obstacles, and turning radius.






\subsection{Evaluation Measures}
In addition to subjective measurements from the NASA questionnaire and the preference questionnaire, we applied four objective metrics to quantify driving performance:

\begin{itemize}
    \item \textit{Jerk of the car movement:} This measures driving motion smoothness and ride comfort by computing the time rate of change of acceleration \cite{schot1978jerk}. 
    \item \textit{Obstacle margin:} This is the minimal distance between the center of the car and the center of the obstacle when passing it, measuring safety while passing obstacles.
    \item \textit{Times off road:} The total number of times the car drives off the road within a trial.
    \item\textit {Off road duration:} The average time duration the car drives off the road in a trial. It serves as another safety measurement to evaluate over-steering scenarios while the driver is avoiding obstacles.
\end{itemize}

\subsection{Statistical Analysis}
As practice and washout trials were designed to stabilize the operator’s performance and eliminate carry-over effects between consecutive driving conditions, we only analyzed the data from the task trials. The results were averaged over five task trials. The data normality was checked using the Shapiro-Wilk test. A two-way repeated measures ANOVA was conducted for normalized data to analyze the effects of speed, haptic feedback, and their interaction. The non-normalized data were adjusted using Aligned Rank Transform (ART) before conducting the ANOVA. Post hoc comparisons between levels of haptic feedback and levels of speed were performed using pairwise t-tests with Bonferroni adjustment. In this study, speed had three levels \{10, 12.5, 15\}\,m/s, and haptic guidance had three stiffness levels \{0, 200, 500\}\,N/m, resulting in a 3×3 design with nine conditions.

\section{Results}
\begin{figure*}[htbp]
\centering
 \includegraphics[width=0.8\textwidth]{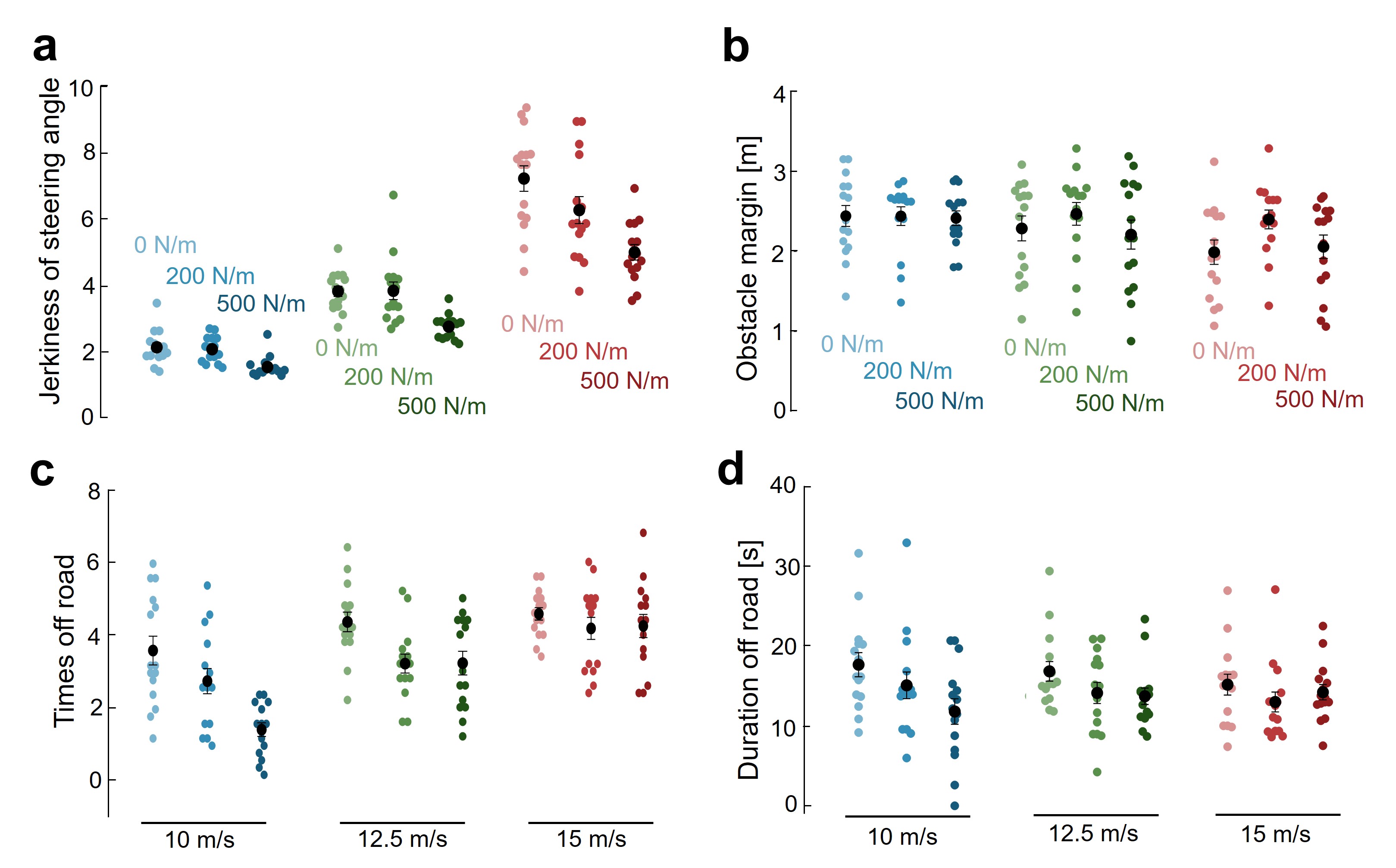}
\caption{Experiment results on a) jerk of steering angle, b) obstacle margin, c) times off the road, and d) duration off the road.}
      \label{fig:results}
\end{figure*}

\subsection{Driving Performance}


\subsubsection{Motion jerk}

Fig.\,\ref{fig:results}a shows the results on rear car's \textit{motion jerk}. There was a statistically significant two-way interaction between speed and haptic feedback ($F(2.36, 33.05) = 7.53, p = 0.001$). Therefore, simple main effects were run. Participants exhibited a similar tendency to jerk across all speed levels for different types of feedback. There was no significant difference in jerk between trials without haptic feedback and those with soft haptic feedback. However, jerk was significantly reduced in trials with large feedback gain compared to those with no feedback ($p < 0.01$) and small feedback gain ($p < 0.01$). Furthermore, in all feedback conditions, lower speeds resulted in less jerk of the steering angle (10 m/s vs. 12.5 m/s, all $p < 0.001$; 12.5 m/s vs. 15 m/s, all $p < 0.001$). This may be due to the fact that at higher speeds, participants have less time to respond to obstacles, leading to sudden changes in steering operations and causing increased jerk in the steering angle.

\subsubsection{Obstacle margin}
The result on the \textit{obstacle margin} is illustrated in Fig.\,\ref{fig:results}b. All conditions are normally distributed (p $>$ 0.08) except the conditions of 10 m/s with small feedback gain (p = 0.001) and 15 m/s with large feedback gain (p = 0.044). There was no statistically significant two-way interaction between speed and haptic feedback (F(4, 56) = 1.01, p = 0.184). The main effect of speed showed a statistically significant difference in obstacle margin (F(2, 28) = 7.981, p = 0.002). The pairwise further showed that the obstacle margin is smaller in trials with a driving speed of 15\,m/s compared to 12.5\,m/s (p = 0.044) and 10\,m/s (p = 0.017), but no statistical difference between 10 m/s and 12.5 m/s (p = 0.316). However, the main effect of haptic feedback did not show a statistical difference in the obstacle margin (F(2,28) = 2.926, p = 0.07).

\subsubsection{Times off the road}

The results on \textit{Times off the road} are normally distributed under all conditions (all p $>$ 0.113). As shown in Fig.\,\ref{fig:results}d. The interaction between speed and haptic feedback is statistically significant (F(4, 56) = 6.575, p $<0.001$). When the driving speed is as slow as 10 m/s, the times off the road are not different from no haptic feedback to small feedback gain (p = 0.054), but large feedback gain has fewer times off the road than small feedback gain ($p < 0.001$) and than no feedback ($p < 0.001$). If the driving speed increases to 12.5 m/s, the trials with both small (p = 0.004) and large (p = 0.008) feedback gains show fewer times off the road than trials with no feedback, there is no difference between small and large feedback gains (p = 1). When the driving speed increased to 15 m/s, there was no observed difference in the times off the road among three levels of haptic feedback F(2,28) = 1.164, p = 0.327.



\subsubsection{Duration off the road}

The results of \textit{Duration off the road} are plotted in Fig.\,\ref{fig:results}d. Almost half of the conditions are not normally distributed. Data were adjusted using ART prior to performing ANOVA \cite{ARTool}. There was no statistically significant two-way interaction between speed and haptic feedback (F(4,56) = 2.02, p = 0.11). The main effect of speed showed no statistical difference in duration off the road (F(2,28) = 0.487, p = 0.625). However, there are significant differences in different levels of haptic feedback (F(2,28) = 7.39, p = 0.003). The trials with haptic feedback are better than those without feedback (small feedback gain vs. no feedback, p = 0.002; large feedback gain vs. no feedback, p = 0.021), but no difference between small and large feedback gains (p = 1). 


\subsection{Workload and Preference}
\begin{figure*}[thpb]
  \centering      
  \includegraphics[width=1\textwidth]{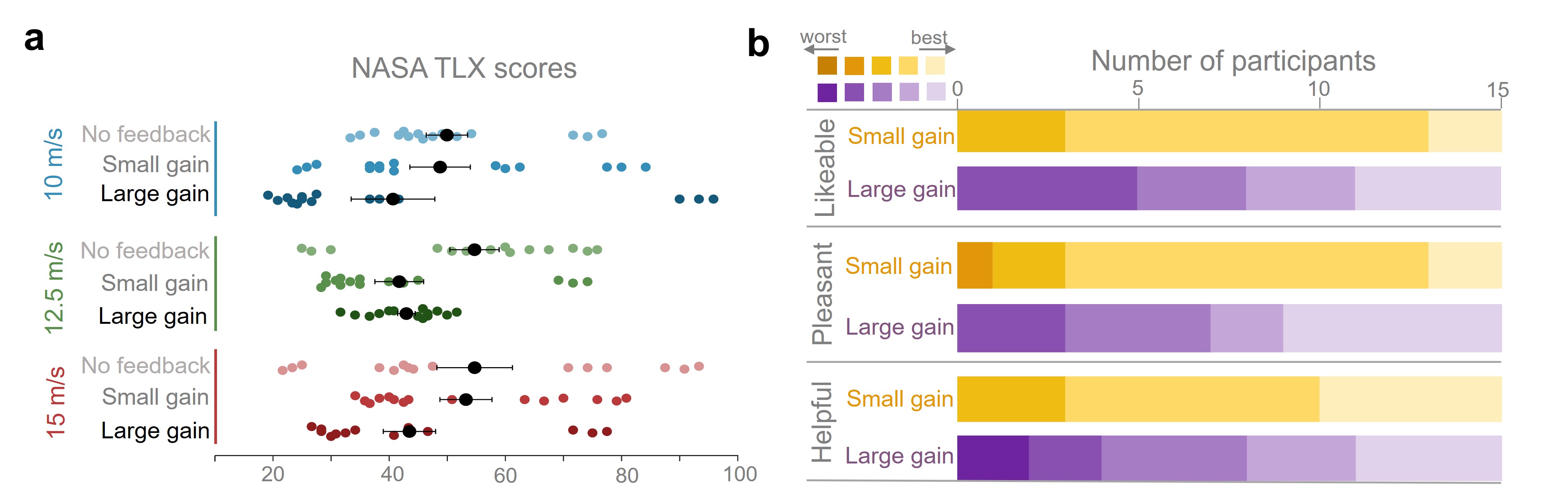}
  \caption{Appreciation on the different conditions. a) NASA Task Load Index result. b) Preference for strong and soft haptic feedback.}
  \label{questionnaire}
\end{figure*}

Fig.\,\ref{questionnaire}a illustrated the NASA TLX questionnaire result on workload. There was no statistically significant two-way interaction between haptic feedback and driving speed (F(4, 56) = 1.17, p = 0.33). We did not observe obvious differences of different speed levels on the workload (F(2, 28) = 2.11, p = 0.14). However, the haptic feedback reduces the workload effectively, the main effect of haptic feedback showed a difference in workload scores ($F(2, 28) = 20.039, p<0.05$). In addition, the post-hoc analysis showed that the workload was marginally reduced from no haptic feedback to small feedback gain (p = 0.055) and was significantly reduced from small feedback gain to large feedback gain (p = 0.007).   



The questionnaire results on subjective haptic experience are shown in Fig.\,\ref{questionnaire}b. In general, most of the participants believed that the haptic feedback provided positive influences. The soft haptic feedback seems preferred compared to the strong haptic feedback, 12/15 participants liked the small feedback gain, while 7/15 liked the large feedback gain. The result of pleasantness and helpfulness is similar, small feedback gain gained more positive responses than large feedback gain.  


After the experiment, most of the participants felt that driving speed was the biggest influence factor on driving performance (5 out of 15 participants ranked driving speed as the first factor, and the other 10 ranked it as the second). In contrast, the participants felt the obstacles affected least on their performance. Similar to the above result, the responses to the haptic feedback varied. 6 out of 15 of the participants believed it affected their performance as the first or second important factor, while the others did not feel its obvious influence.  

\section{Discussion}
Studies from the last decade have shown that individuals connected by a virtual spring spontaneously share their motion plans through this haptic channel, which they can use to improve their motion control \cite{Ganesh2014, Takagi2017, takagi2019individuals}. This \textit{sensory augmentation} relies on each agent's sensing abilities, the mechanical connection between them to transfer haptic information, and their capacity to form a motion plan for their partner(s) during a common task. Given that cars moving one after the other on a road share the same transport task, transferring motion information from a frontal car to a following car's driver using a virtual mechanical connection could potentially improve vehicle control.
 
We tested this hypothesis by developing a shared-haptic virtual platform to simulate a scenario where a front car is driven by an MPC-simulated expert driver and a rear car is driven by a human driver receiving haptic feedback of the front car’s motion through a virtual mechanical connection. A user experiment was carried out using this virtual driving system to study how haptic feedback reflecting the front car’s motion influences driving performance metrics at various speeds and with randomly placed obstacles. Previous studies on haptic-shared driving have focused on single-car scenarios, in contrast to our approach, which uses haptic feedback to convey the motion of surrounding cars.
 
The results of our experiment showed that haptic feedback improves the human driver’s control performance in the rear car. Specifically, the car's motion jerk decreased, particularly in the fastest speed group when strong haptic feedback was provided. The duration and frequency of the car driving off the road were reduced, likely because the haptic feedback incorporating the front car's motion prevented over-reactions from the rear car driver while avoiding obstacles. The obstacle margin remained at a similar safe level with haptic feedback, as the driver had sufficient time to plan their control and avoid obstacles. The questionnaire responses support that haptic feedback effectively reduced mental load, and most participants preferred receiving haptic information about the front car's motion.
 
The optimal level of haptic feedback depends on the specific driver and scenario. While there was an overall preference for soft haptic feedback, some participants preferred strong haptic feedback, possibly because the automated controller offered superior control, or because driving took too much time and effort. Adaptive haptic feedback may be effective, with levels increasing in high-velocity and complex scenarios where automation excels, and decreasing in gentler scenarios that the human driver can handle well.
 
The experimental conditions had different impacted the evaluation metrics. Driving speed significantly influenced ride comfort and car controllability. Specifically, motion jerk increased with driving speed, while obstacle margin decreased. Our results show that the potential risks associated with increasing driving speeds could be effectively mitigated by haptic feedback. This suggests the potential for tailoring haptic feedback levels for different transportation scenarios based on their specific purposes.
 
Future testing of our hypothesis should use a more realistic driving simulator with immersive 3D visualization to study the effects of haptic communication and its interaction with other driving factors, such as driver view or road conditions. Additionally, we used an autonomous car as the leading car to produce consistent motion and simulate an expert driver for the following car's driver. It would be interesting to evaluate the robustness of these results with varied driving behaviors controlling the leading car.


%

\section*{ACKNOWLEDGMENT}
We thank Dr. Cecilia De Vicariis for her help in the experiment setup, and all the subjects for participating in the experiment. 


\bibliographystyle{IEEEtran}
\bibliography{IEEEabrv,Reference}


\end{document}